\documentclass{article}
\usepackage{spconf,amsmath,graphicx,cite}
\usepackage{siunitx,wrapfig}

\title{In-Sensor \& Neuromorphic Computing are all you need for Energy Efficient Computer Vision}
\name{\begin{tabular}{c}Gourav Datta$^1$, Zeyu Liu$^1$, Md Abdullah-Al Kaiser$^{1,2}$, Souvik Kundu$^3$, Joe Mathai$^2$, \\ Zihan Yin$^{1,2}$, Ajey P. Jacob$^2$, Akhilesh R. Jaiswal$^{1,2}$, Peter A. Beerel$^1$\end{tabular}}
\address{$^1$University of Southern California, USA \ \ $^2$Information Sciences Institute, USA \ \ $^3$Intel Labs, USA}
\begin{document}
%
\maketitle
\begin{abstract}
Due to the high activation sparsity and
use of accumulates (AC) instead of expensive multiply-and-accumulates (MAC), neuromorphic \textit{spiking neural networks} (SNNs) have emerged as a promising low-power alternative to traditional DNNs for several computer vision (CV) applications. However, most existing SNNs require multiple time steps for acceptable inference accuracy, hindering real-time deployment and increasing spiking activity and, consequently, energy consumption. Recent works proposed direct encoding that directly feeds the analog pixel values in the first layer of the SNN in order to significantly reduce the number of time steps. Although the overhead for the first layer MACs with direct encoding is negligible for deep SNNs and the CV processing is efficient using SNNs, the data transfer between the image sensors and the downstream processing costs significant bandwidth and may dominate the total energy. To mitigate this concern, we propose an \textit{in-sensor computing} hardware-software co-design framework for SNNs targeting image recognition tasks. Our approach reduces the bandwidth between sensing and processing by $12{-}96\times$ and the resulting total energy by $2.32\times$ compared to traditional CV processing, with a $3.8\%$ reduction in accuracy on ImageNet.

\end{abstract}
%
%
\vspace{-2mm}
\section{Introduction}
\label{sec:intro}
\vspace{-1mm}
The demand to process vast amounts of data generated from the high frame-rate, high-resolution cameras has motivated novel energy-efficient on-device CV solutions \cite{pinkhan2021jetcas,scamp2020eccv,datta2022scireports}. Visual data in such cameras are usually captured in analog voltages by a sensor pixel array and then converted to the digital domain for subsequent AI processing using analog-to-digital converters (ADC). 
Hence, high-resolution input images need to be streamed  between the camera and the CV processing unit, frame by frame, causing energy, bandwidth, and security bottlenecks \cite{datta2022scireports}. In fact, this energy may actually dominate the total energy incurred by the CV processing, particularly with the range of effective model optimization techniques developed by the CV community \cite{model_compression}. 
One such technique that we adopt in this work is the conversion of the traditional networks to one-time-step SNNs \cite{datta2022onestep} (no temporal overhead) that achieve high sparsity and avoid expensive multiplication operations, thereby improving energy efficiency \cite{neuro_frontiers,dsnn_conversion1}.

To mitigate the data transfer bottleneck, we propose an in-sensor computing paradigm that customizes the pixel array and periphery to implement the analog multi-channel, multi-bit convolution required in direct encoding, batch normalization (BN), and comparison operation required for leaky-integrate-and-fire (LIF) SNN models. Our solution includes a holistic algorithm-circuit co-design approach and achieves bandwidth reduction via the reduction from direct encoded full-precision input images to spiking output feature maps obtained via large strides and reduced channels with knowledge distillation (KD). Compared to existing in-sensor computing approaches for CNNs, we present three distinct contributions for our one-time-step SNNs. (1) We re-purpose the analog \textit{correlated double sampling} (CDS) circuit present in image sensors to implement positive and negative weights in the first SNN convolutional layer. (2) We propose a simple 2T-based analog comparator circuit to process the SNN LIF layer whose trip point is adjusted such that the BN layer can be fused in the preceding convolutional and succeeding LIF layer without additional hardware. (3) We train our SNNs using a variant of KD to reduce the \# of channels in the in-sensor convolution for reduced bandwidth and data transfer cost.  

\begin{figure*}
\begin{center}
\includegraphics[width=0.8\textwidth]{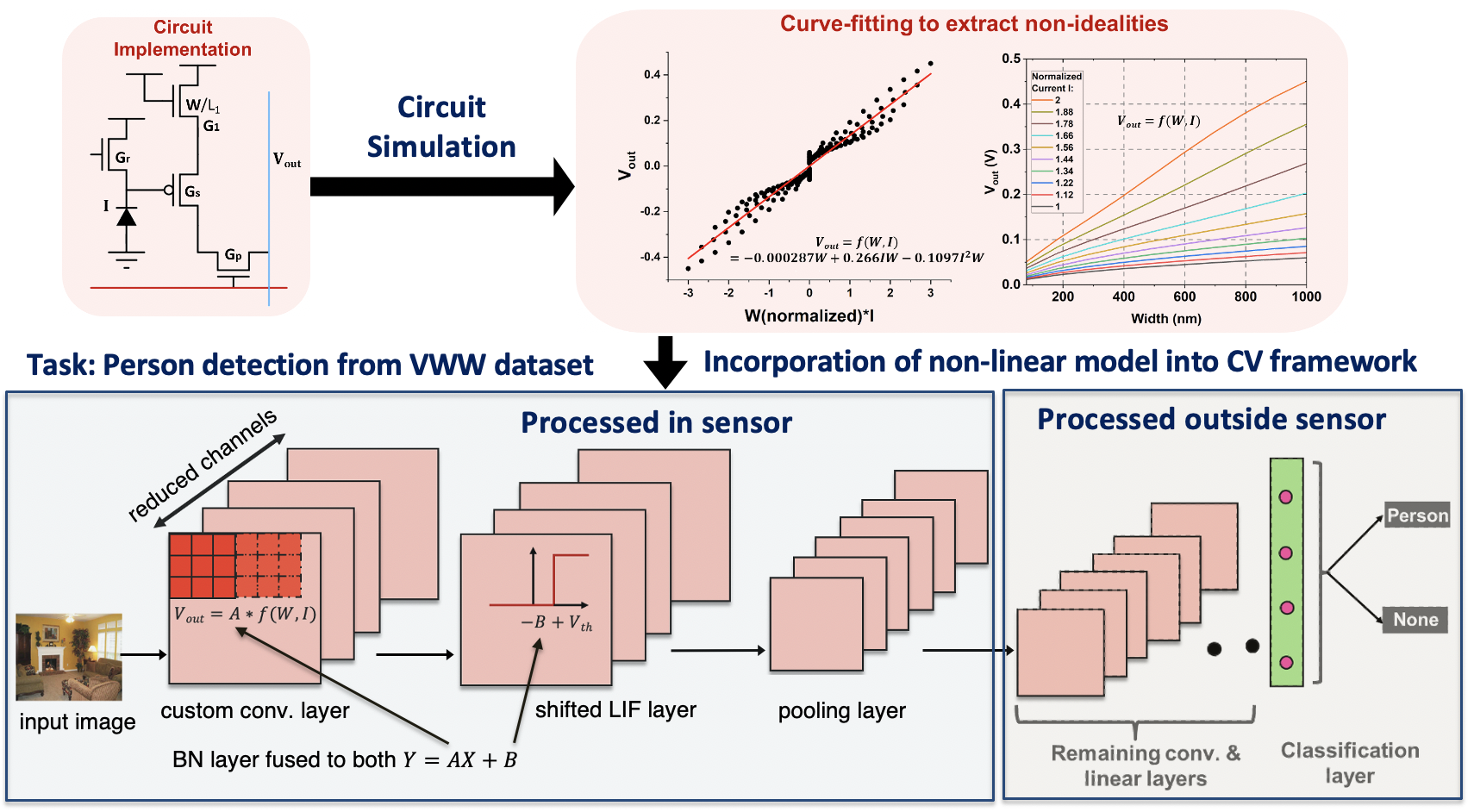}
\end{center}
\vspace{-3mm}
\caption{Proposed in-sensor \& neuromorphic computing framework for energy- \& bandwidth-efficient deep learning.}
\vspace{-2mm}
\label{fig:proposed_framework}
\end{figure*}

\vspace{-2mm}
\section{Related Work}\label{sec:background}
\vspace{-2mm}
\subsection{Spiking Neural Networks}

Since SNNs receive and transmit information via spikes, analog inputs have to be encoded with a sequence of spikes using one of several known techniques including rate coding \cite{diehl2016conversion}, temporal coding \cite{comsa_2020}, and rank-order coding \cite{Kheradpisheh_2018}. 
In this work, we adopt the recently proposed direct encoding technique \cite{rathi2020dietsnn, datta2021deep,datta2021ijcnn,spiking_lstm} that directly feeds the analog pixel values into the first layer and emits spikes only in the subsequent layers. 
Although the first layer now requires MACs, the overhead is negligible for deep CNNs \cite{rathi2020dietsnn,datta2022fin} and the number of time steps needed to achieve state-of-the-art (SOTA) accuracy reduces significantly. Coupled with conversion from DNNs by approximating the activation value of ReLU neurons with the firing rate of spiking
neurons \cite{dsnn_conversion_abhronilfin,dsnn_conversion1}, supervised learning algorithms for SNNs have overcome various roadblocks associated with the discontinuous spike activation function \cite{lee_dsnn,kim_2020,kundu2021lowlatency,Kundu_2021_WACV}. In addition to these techniques, we leverage our recently proposed SNN training technique using Hoyer regularizer \cite{datta2022onestep} that yields ultra-sparse one-time-step models with significant improvemenets energy efficiency compared to SOTA.

\vspace{-2mm}

\subsection{In-Sensor Computing}

To address the data transfer bottleneck between sensors and accelerators, researchers have proposed several near-sensor \cite{pinkhan2021jetcas} and in-sensor \cite{chen2020pns,scamp2020eccv} processing approaches. Near-sensor processing aims to incorporate a dedicated DL accelerator chip on the same printed circuit board \cite{pinkhan2021jetcas}, or even 3D-stacked with the CMOS image sensor chip.
On the other hand, in-sensor processing integrates digital or analog circuits embedded within either the individual pixels \cite{scamp2020eccv} or the periphery of the sensor chip \cite{chen2020pns}, reducing the data transfer between the sensor and CV processing chips. 
However, most existing approaches fail to support multi-bit, multi-channel convolution, BN, and ReLU operations needed for complex CV tasks, and sometimes require emerging materials, that are incompatible with foundry-manufacturing of image sensors.
These concerns are addressed in our recent works \cite{datta2022scireports,detrack,datta2022hsipip} where the weights and input activations are available within individual pixels.

\vspace{-2mm}
\section{Overview and Challenges}
\vspace{-1mm}
We propose to implement the first convolutional layer of our SNN model by embedding appropriate weights inside pixels which includes the spatial and output channel dimensions \cite{datta2022scireports} as shown in Fig. \ref{fig:proposed_framework}. The convolutional weights are encoded as transistor widths which are fixed during manufacturing \cite{datta2022scireports}. Note that this lack of programmability is not a significant problem because the first few layers of our SNN model (similar to CNNs) extract high-level spectral features that can be common across various benchmarks. The weights can also be programmable by mapping to emerging resistive non-volatile memory elements embedded within individual pixels.
By activating multiple pixels simultaneously, the weight-modulated outputs of different pixels are summed together in parallel in the analog domain, effectively performing a convolution operation. In order to capture the non-idealities of our in-sensor convolution, we simulate the pixel output voltage with varying weights and inputs, the latter reflecting the photo-diode currents (similar to \cite{datta2022scireports}). We then use standard curve-fitting to generate a custom CNN layer that replaces the element-wise multiplication in standard convolution. 

Such in-sensor computing is challenging because image sensors are optimized for reading out pixels and not processing computational requirements of modern CNN/SNN layers such as multi-bit, multi-channel strided convolution, batch-normalization (BN), ReLU/LIF, and pooling. This requires addressing tightly-intertwined algorithm-hardware co-design challenges because (i) the transistor widths encoding the in-sensor convolutional weights can only be positive, whereas, CNNs require both positive and negative weights, particularly in the initial layers for SOTA accuracy, and (ii) hardware manufacturing constraints motivate low-bit-width, large stride, and reduced \# of channels which directly conflict with the algorithmic accuracy requirement.
\vspace{-2mm}
\section{Proposed Co-Design Framework}
\label{sec:proposed_framework}
\vspace{-1mm}
Our proposed hardware-algorithm co-design framework enables efficient computer vision by (1) leveraging SOTA SNN training techniques to reduce the energy consumption of the vision processing, and (2) embedding computations of the first few layers of such models in the image sensor with manufacturing-friendly hardware platforms to reduce the sensing, ADC, and data transfer energy. This reduction is obtained from the compressing the activation maps coming from the sensor primarily as a result of (a) 1-bit activation in SNNs compared to multi-bit pixels from image sensors, and (b) effective reduction  of the \# of channels in the first layers using a variant of KD, and (c) spatial dimension reductions from strided convolution and pooling. 

Compared to existing in-sensor computing approaches that implement traditional CNNs \cite{datta2022hsipip,datta2022scireports}, our proposal improves the hardware efficiency since we replace the expensive ADC operations (for ReLU) with simple comparison operations (for LIF/IF) that yield a 1-bit output. 
However, the ADC in existing works \cite{datta2022scireports} also realizes both the positive and negative weights using the differential sensing nature of its digital CDS circuit. We propose to mitigate this concern by adopting analog CDS circuits present in traditional image sensors. Additionally, the ADC in existing works helps fuse the BN layer partly in the ReLU layer (and partly in the convolution) by resetting the counter to a non-zero value \cite{datta2022scireports}. We propose to bridge this gap for SNNs by tuning the threshold of the analog comparison operation. The details of these approaches along with the SNN training algorithm are described below.      

\begin{figure}
    \centering
    \includegraphics[width=0.4\textwidth]{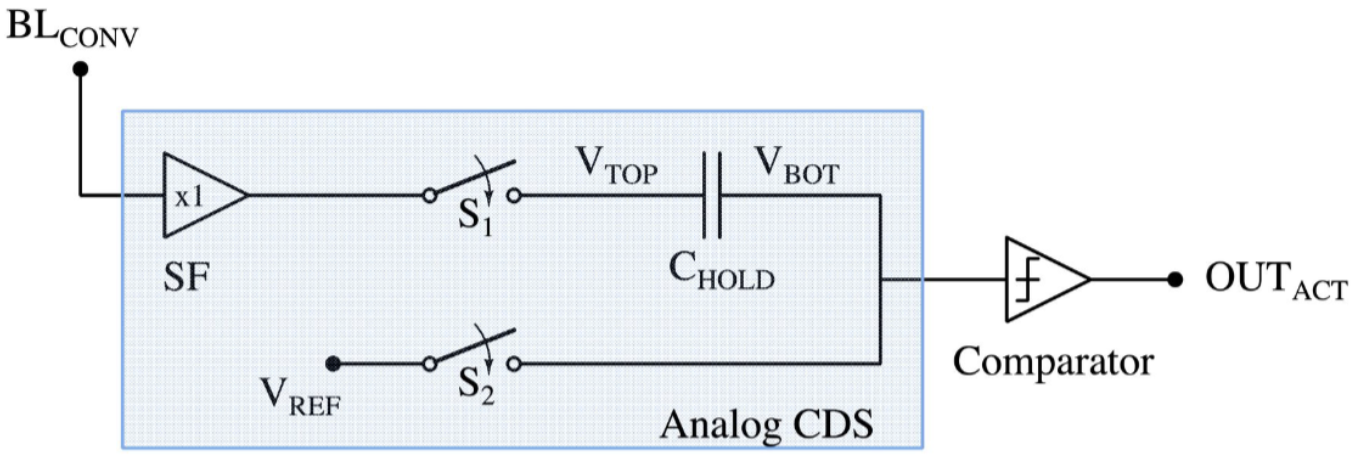}
    \caption{Proposed analog CDS and comparator circuits to realize SNN-based in-sensor computing.}
    \label{fig:analog_CDS_comparator}
    \vspace{-5mm}
\end{figure}
\vspace{-1mm}

\subsection{Proposed circuits}

Our approach re-purposes the existing on-chip column-parallel analog CDS circuit \cite{analog_CDS} of traditional image sensors to accumulate the convolution output for both negative and positive weights sequentially in two phases, thereby resulting in a subtraction operation \cite{analog_sub} as shown in Fig. \ref{fig:analog_CDS_comparator}. In Phase-I, we sample the accumulation voltage (\si{BL_{CONV}}) corresponding to the negative weights on the top capacitor plate (\si{V_{TOP}}) and a constant reference voltage on the bottom plate (\si{V_{BOT}}) of the holding capacitor (\si{C_{HOLD}}). In Phase-II, the bottom plate is kept floating, while the top plate samples the accumulation voltage (\si{BL_{CONV}}) of the positive weights. As a result, the floating bottom plate of the capacitor follows the difference voltage of the top plate from a constant DC reference voltage (\si{V_{REF}}). The difference between the accumulated voltages calculated by the analog subtractor is compared with a threshold voltage using an inverted-based 2T comparator (see Fig. \ref{fig:analog_CDS_comparator}). The trip point (threshold voltage) of the comparator can be skewed by modulating the bias voltage of the pMOS transistor of the comparator.  When the difference in accumulation voltages is higher (lower) than the comparator's threshold, the comparator generates a high (low) output activation (\si{O_{ACT}}). This \si{O_{ACT}} is then fed to the succeeding pooling layer which performs a 'max' operation for each kernel in the periphery of the sensor.
\vspace{-3mm}

\subsection{Fusion of BN layer}

Consider a BN layer (that is critical to converging one-time-step SNNs) with $\gamma$ and $\beta$ as the trainable parameters, and a running mean and variance of $\mu$ and $\sigma$ respectively, which are saved and used for inference. Such a BN layer implements a linear function during inference, as shown below.
\vspace{-1mm}
\begin{equation}
Y=\gamma\frac{X-\mu}{\sqrt{\sigma^2+\epsilon}}+\beta=\underbrace{\left(\frac{\gamma}{\sqrt{\sigma^2+\epsilon}}\right)}_{A}\cdot X+\underbrace{\left(\beta-\frac{\gamma\mu}{\sqrt{\sigma^2+\epsilon}}\right)}_{B}  \notag
\end{equation}
\noindent
We fuse the scale term $A$ into the preceding convolutional layer weights (value of the pixel embedded weight tensor is $A\cdot\theta$, where $\theta$ is the final weight tensor obtained by our training) that are embedded as the transistor widths. Additionally, we propose to shift the trip point of the analog comparator, as shown in Fig. \ref{fig:proposed_framework} to incorporate the shift term $B$ i.e., assuming a LIF threshold voltage of $V_{th}$ obtained from SNN training, the trip point becomes ${-}B{+}V^{th}$. 
\vspace{-2mm}
\subsection{Training SNN model with bandwidth reduction goals}

We train a one-time-step SNN model using the recently proposed variant of the Hoyer regularizer that reduces the energy consumption of the CV processing by an order of magnitude compared to non-spiking CNNs. This implies that the image sensor energy and the data transfer energy from the sensor to the CV processing unit might be crucial to be minimized. We achieve this goal using our proposed algorithm-hardware co-design approach detailed above. 
To further reduce these energies, we attempt to reduce the \# of output channels in the in-sensor convolutional layer by distilling knowledge from a teacher SNN model with a large \# of channels during training. In particular, we apply a $1{\times}1$ convolution on the in-sensor convolutional output to match the activation map dimension with the teacher model. In addition to the standard cross-entropy loss, we also incorporate the difference of the logits before softmax activation at the output \cite{ba2014deep}, and the L2 difference of the activation map \cite{komodakis2017paying} between the teacher and student models in our loss function. Based on empirical evidence of accuracy gain, we incorporate the difference in the output activations of other basic blocks in our loss function.

\section{Experimental Results}
\label{sec:exp}

We evaluate our proposed framework against a baseline where the entire SNN is processed outside the sensor on object recognition tasks from CIFAR10 \cite{cifar10} and ImageNet \cite{imagenet_cvpr09} datasets using variants of VGG \cite{vgg} and ResNet \cite{resnet} architectures. We use similar training hyperparameters as \cite{datta2022onestep}. 

\begin{table}[t]
\caption{Comparison of the test accuracy of our in-sensor and baseline one-time-step SNN models w \& w/o our KD.} 
\vspace{-4mm}
\label{tab:results}
\begin{center}
\setlength{\tabcolsep}{.5mm}{
\begin{tabular}{l|c|c|c|c|c}
\hline
\hline
Dataset & Network & Imple- & Channel & Acc (\%) & Acc (\%)\\
 &  & mentation  & count & & (with KD)  \\
\hline
CIFAR10 & VGG16 & Baseline & 64  & 93.16 & - \\
\hline
CIFAR10 & VGG16 & In-sensor & 64  &  91.85 & - \\
\hline
CIFAR10 & VGG16 & In-sensor & 16  &  90.56 & 91.27 \\
\hline
CIFAR10 & VGG16 & In-sensor & 8 &  89.12 & 89.38 \\
\hline
CIFAR10 & ResNet20 & In-sensor & 64 & 91.38  & - \\
\hline
CIFAR10 & ResNet20 & In-sensor & 8 &  89.69 & 90.20 \\
\hline
ImageNet & VGG16 & Baseline & 64  & 66.00 & - \\
\hline
ImageNet & VGG16 & In-sensor & 64 &  63.82 & - \\
\hline
ImageNet & VGG16 & In-sensor & 16 &  59.11 & 60.02 \\
\hline

\end{tabular}
}
\vspace{-4mm}
\end{center}
\end{table}
\vspace{-3mm}
\subsection{Accuracy}

As illustrated in Table \ref{tab:results}, our in-sensor SNN models yield accuracies within $1.31\%$ and $2.18\%$ of the baseline models on CIFAR10 and ImageNet respectively. The VGG and ResNet based models incur similar loss in accuracy. This loss is actually due to the custom convolutional layer with reduced representation capacity due to hardware constraints and non-idealities. However, reducing the in-sensor convolutional channels for bandwidth reduction goals by $4\times$ degrades the accuracy by $1.29\%$ on CIFAR10 and $4.71\%$ on ImageNet. Our KD technique helps reduce the accuracy gap to $0.58\%$ on CIFAR10 and $3.80\%$ on ImageNet.

\begin{figure}
    \centering
    \includegraphics[width=0.4\textwidth]{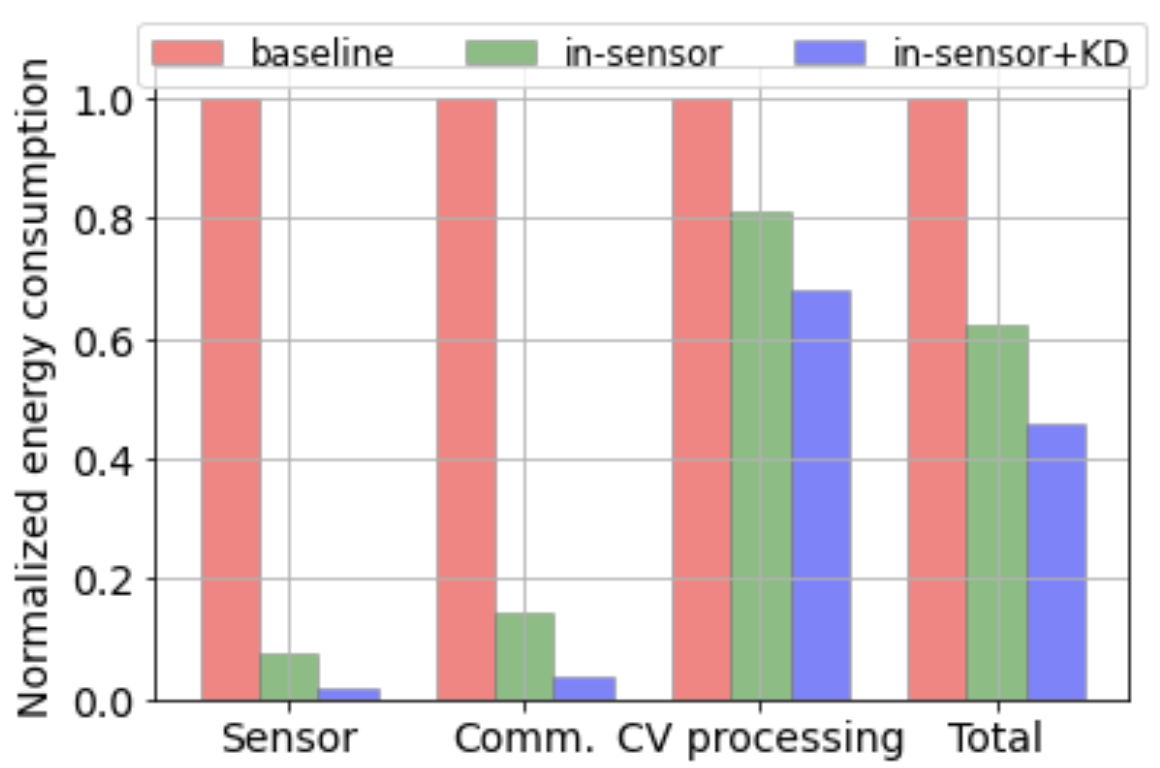}
    \caption{Comparison of different energy components between the baseline and proposed in-sensor computing approaches.}
    \label{fig:energy_results}
    \vspace{-3mm}
\end{figure}
\vspace{-2mm}

\subsection{Energy \& Bandwidth reduction}

The bandwidth reduction obtained by our framework can be computed as 
\begin{equation}\label{eq:pip_compression}
    C = \left(\frac{{H^o{\times}{W^o}{\times}{C^o}}}{{H^i{\times}{W^i}{\times}{C^i}}}\right)\cdot\frac{N_{im}}{N_{sp}}\cdot\frac{4}{3}
\end{equation}
where $H$, $W$, $C$ denote the height, width, \#  of channels respectively and the superscript $i$ and $o$ denote the image input and in-sensor output (after convolutional, BN, LIF, and pooling layers) respectively. $N_{im}$ denotes the pixel bit-precision (${\sim}12$ in image sensors \cite{onsemi:AR0135AT}), and $N_{sp}{=}1$ denotes the in-sensor output bit-precision. The factor $\left(\frac{4}{3}\right)$ represents the compression from Bayer’s pattern of RGGB pixels to RGB pixels. 
Plugging these values for VGG16, we obtain $C{=}12$. Reducing the \# of in-sensor conv. channels by $4{-}8\times$ (see Table 1) via KD increases $C$ by the same factor (upto $96\times$).

We calculate the total energy consumption of our baseline (processed completely outside the pixel array) and in-sensor SNN models for VGG16 on ImageNet. 
The total energy is computed as the sum of the image sensor energy, the sensor-to-SoC communication energy obtained from \cite{lvds}, and the energy to process the SNN layers outside the sensor. The sensor energy is the sum of the image read-out energy and the ADC energy for the baseline models. For our in-sensor models, it is the sum of the pixel convolution energy, analog CDS, and comparator energy. All these energy values are obtained from in-house circuit simulations in GF22FDX technology, except the ADC energy that is obtained from \cite{gonugondla2021imc}. We use the analytical model used in \cite{kang2018imvlsi,datta2021deep} to compute the energy required to process the SNN layers (total $L$) for the baseline and in-sensor models.
This model is a function of the energy incurred in reading each element from the on-chip memory to the processing unit, and the energy consumed in each MAC (AC) operation. Note that the direct encoded SNN receives analog input in the first layer ($l{=}1$), thereby incurring MACs without any sparsity \cite{datta2022onestep,rathi2020dietsnn}. Their values are obtained from \cite{kang2018imvlsi}, applying voltage scaling for iso-V$_{dd}$ conditions with other energy estimations.

     

As shown in Fig. \ref{fig:energy_results}, our in-sensor SNN model without KD yields $13.3\times$, $6.9\times$, and $1.23\times$ reduction in the sensor energy, data transfer energy, and off-chip energy respectively compared to the baseline counterpart. 
Reducing the \# of channels by $4\times$ with KD increases the gains of these three energy components to $54.1\times$, $27.8\times$ and $1.47\times$ respectively. This leads to a total energy reduction of $2.17$ which can be further increased to $3.56$ using a more efficient backbone, MobileNetV2. Compared to traditional CNN-based in-sensor computing, our approach yields a $11.2\times$ reduction in total energy with VGG16 on ImageNet.     
\vspace{-2mm}
\section{Conclusions}

This work proposes in-sensor computing based algorithm-hardware co-design framework that can 
reduce the energy consumption of a complex object recognition pipeline by up to $2.32\times$ compared to traditional vision processing approaches. Our hardware modifications can be readily integrated into the foundry-manufacturable CMOS image sensor platforms, and our hardware-inspired algorithmic modifications are shown to yield negligible performance drop in complex vision tasks. 
Note that the energy benefits reported in this work is based on the assumption that the front-end sensor and vision processing chips are closely located on the same printed-circuit board \cite{kodukula2020sensors}. In general, they could be separated by large distances, necessitating long energy-expensive wired or wireless data transfer (which is common for the case of sensor-fusion and swarm intelligence applications). In such cases the overall energy improvement would approach the bandwidth reduction (up to 96$\times$).

\bibliographystyle{IEEEbib}
\bibliography{strings,refs}

\end{document}